\documentclass{article}

\usepackage[final]{neurips_2020_ml4ps}

\usepackage[utf8]{inputenc} 
\usepackage[T1]{fontenc}    
\usepackage{url}            
\usepackage{booktabs}       
\usepackage{amsfonts}       
\usepackage{nicefrac}       
\usepackage{microtype}      
\usepackage{bm}
\usepackage{enumitem}
\usepackage{graphicx}
\usepackage{amsmath}
\usepackage{subfigure}
\usepackage{algorithm}
\usepackage{algorithmic}
\usepackage[numbers, sort]{natbib}

\usepackage[colorlinks=true,bookmarksopen,pdfsubject={algorithms},linkcolor={blue}, anchorcolor={black}, citecolor={blue}, filecolor={magenta}, menucolor={black}, pagecolor={red},backref=none,urlcolor={blue}]{hyperref}
\title{A Hybrid Gradient Method to Designing Bayesian Experiments for Implicit Models}

\author{Jiaxin Zhang \\
  Computer Science and Mathematics Division\\
  Oak Ridge National Laboratory, Oak Ridge, TN 37830 \\
  \texttt{zhangj@ornl.gov} \\
   \And
   Sirui Bi \\
   Computational Sciences and Engineering Division  \\
   Oak Ridge National Laboratory, Oak Ridge, TN 37830 \\
  \texttt{bis1@ornl.gov} \\
   \AND
   Guannan Zhang \\
   Computer Science and Mathematics Division \\
   Oak Ridge National Laboratory, Oak Ridge, TN 37830 \\
  \texttt{zhangg@ornl.gov} \\
}

\begin{document}

\maketitle

\begin{abstract}
Bayesian experimental design (BED) aims at designing an experiment 
to maximize the information gathering from the collected data. The optimal design is usually achieved by maximizing the mutual information (MI) between the data and the model parameters. When the analytical expression of the MI is unavailable, e.g., having implicit models with intractable data distributions, a neural network-based lower bound of the MI was recently proposed and a gradient ascent method was used to maximize the lower bound \cite{kleinegesse2020bayesian}. However, the approach in \cite{kleinegesse2020bayesian} requires a pathwise sampling path to compute the gradient of the MI lower bound with respect to the design variables, and such a pathwise sampling path is usually inaccessible for implicit models. In this work, we propose a hybrid gradient approach that leverages recent advances in variational MI estimator and evolution strategies (ES) combined with black-box stochastic gradient ascent (SGA) to maximize the MI lower bound. This allows the design process to be achieved through a unified scalable procedure for implicit models without sampling path gradients. Several experiments demonstrate that our approach significantly improves the scalability of BED for implicit models in high-dimensional design space. 


\end{abstract}

\section{Introduction}

Experimental design plays an essential role in all scientific disciplines. Our ultimate goal is to determine designs that maximize the information gathered through the experiments so that improve our understanding on model comparison or parameter estimations. A broadly used approach is Bayesian experimental design (BED) \cite{chaloner1995bayesian} that aims to find an optimal design $\bm{\xi}^*$ to maximize a utility function $I(\bm \xi)$, which is typically defined by the mutual information (MI) between data and model parameters. Typically, the BED framework begins with a Bayesian model of the experimental process, including a prior distribution $p(\bm \theta)$ and a likelihood $p(\bm y|\bm \theta,\bm \xi)$. The information gained about $\bm \theta$ from running the experiment with design $\bm \xi$ and observed outcome $\bm y$ can be interpreted by the reduction in entropy from the  prior to posterior 
\begin{equation}
    \rm IG(\bm y, \bm \xi) = \mathcal{Q}[p(\bm \theta)] - \mathcal{Q}[p(\bm \theta | \bm y, \bm \xi)].
\end{equation}
To define a metric to quantify the utility of the design $\bm \xi$ before running experiments, an expected information gain (EIG), $I(\bm \xi)$ is often used:
\begin{equation}
    I(\bm \xi) = \mathbb{E}_{p(\bm y|\bm \xi)}[\mathcal{Q}[p(\bm \theta)] - \mathcal{Q}[p(\bm \theta | \bm y, \bm \xi)]]. \label{eq:eig}
\end{equation}
Eq.~\eqref{eq:eig} can also be interpreted as a mutual information (MI) between $\bm \theta$ and $\bm y$ with a specified $\bm \xi$, 
\begin{equation}
    I_{\rm MI} (\bm \xi) = \mathbb{E}_{p(\bm \theta)p(\bm y|\bm \theta, \bm \xi)} \left[ \log\frac{p(\bm y|\bm \theta, \bm \xi)}{p(\bm y | \bm \xi)} \right] \label{eq:BED_MI}
\end{equation}
The Bayesian optimal design is therefore defined as 
\begin{equation}
    \bm \xi^* = \operatorname*{arg\,max}_{\bm \xi \in \bm \Xi} I_{\rm MI} (\bm \xi) \label{eq:max_mi}
\end{equation}
where $\bm \Xi$ is the feasible design domain. The most challenging task in BED framework is how to efficiently and accurately estimate $I_{\rm MI} (\bm \xi)$ in Eq.~\eqref{eq:BED_MI} and optimize $I_{\rm MI} (\bm \xi)$ via Eq.~\eqref{eq:max_mi} to obtain the optimal design $\bm \xi^*$. 

Most of the existing BED studies focus on the {\em explicit models} \cite{chaloner1995bayesian, sebastiani2000maximum, foster2019variational,foster2020unified} in which the likelihood is analytically known, but in natural and physical science, a more common scenario is the {\em implicit models} \cite{kleinegesse2019efficient,overstall2018bayesian}, in which the likelihood is intractable but sampling is possible. In other words, the implicit model is specified based on a stochastic data generating simulator and typically has no access to the analytical form and the gradients of the joint density $p(\bm{\theta},\bm{y}|\bm \xi)$ and marginal density $p(\bm y|\bm \xi)$. The resulting BED scheme shares a two-stage feature: build a pointwise estimator of $I(\bm \xi)$ and then feed this ``black-box" estimator to a separate outer-level optimizer such as Bayesian optimization to find the optimal design $\bm{\xi}^*$. This scheme substantially increases the overall computational cost and is challenging in scaling the BED to a high dimensional design space. 

Recent studies \cite{kleinegesse2020bayesian, foster2020unified,harbisher2019bayesian} alleviate the challenges by using stochastic gradient-based approaches but they rely on the models with tractable likelihood functions or assume the gradients can be reasonably approximated by pathwise gradient estimators with sampling path, unlike the scope of our paper that focuses on the BED for implicit models without gradients. We develop a general scalable framework that can jointly optimize a unified objective with respect to both the variational MI bound and the design using a hybrid gradient ascent approach. The key contributions are summarized as follows: 
\vspace{-0.2cm}
\begin{itemize}[leftmargin=10pt]
\item We propose a hybrid gradient approach that leverages recent advances in variational MI estimator and guided evolution strategies (ES) to maximize the MI lower bound;
    \item We incorporate a smoothed estimator of the MI lower bound via clipped density ratios to reduce the variance in the MI bound estimator and the estimation of the posterior samples;
    \item We demonstrate the superior performance of our proposed approach on a toy noisy linear problem and a real quantum control problem, specifically in high dimensional design spaces.  
\end{itemize}

\section{Hybrid Gradient Method for BED}
Here, we aim to find the optimal design $\bm{\xi}^*$ by maximizing a MI lower bound rather than exactly estimating MI with high accuracy. Estimating and optimizing MI is core to many machine learning research but it has been a challenge to bounding MI in high dimensions. 

{\bf Mutual information estimators} \quad Belghazi et al. \cite{belghazi2018mine} proposed to estimate the MI using gradient descent over neural networks and argued that the lower bound can be tightened by optimizing the neural network parameters. The MI estimator is typically named by MINE-$f$ or $f$-GAN KL \cite{nowozin2016f}
\begin{equation}
    I_{\rm MINE}(\bm \xi, \bm \psi) = \mathbb{E}_{p(\bm{\theta},\bm{y}|\bm \xi)}[\mathcal{T}_{\bm \psi}(\bm \theta, \bm y)] - \log \mathbb{E}_{p(\bm{\theta})p(\bm y|\bm \xi)}[e^{\mathcal{T}_{\bm \psi}(\bm \theta, \bm y)}] \label{eq:mine}
\end{equation}
where $\mathcal{T}_{\bm \psi}(\bm \theta, \bm y)$ is a neural network that is parametrized by $\bm \psi$ with model parameters $\bm \theta$ and data $\bm y$ as inputs. Incorporating neural network parameters $\bm \psi$ with design parameters $\bm \xi$, the BED problem can be formulated by maximizing the overall objective
\begin{equation}
    \bm{\xi}^* = \operatorname*{arg\,max}_{\bm \xi}\max_{\bm \psi} \left\{I_{\rm MINE}(\bm \xi, \bm \psi) \right\}. \label{eq:obj}
\end{equation}
The optimal design $\bm{\xi}^*$ is obtained by maximizing the MI estimator in Eq.~\eqref{eq:mine} through a joint gradient-based algorithm or a separate gradient-free update scheme of $\bm \xi$ and $\bm \psi$. The effectiveness for the MI estimation and optimization therefore becomes very important to the BED problem. Unfortunately, $I_{\rm MINE}$ exhibits a high variance that could grows exponentially with the ground truth MI and leads to poor bias-variance trade-offs in practice \cite{poole2019variational,song2019understanding}. To address the high-variance issue in the $I_{\rm MINE}$ estimator, we propose to use a smoothed MI lower-bound estimator $I_{\rm SMILE}$ \cite{song2019understanding} with hyperparameter $\tau$ that clips the density ratios when estimating the partition function: 
\begin{equation}
    I_{\rm SMILE}(\bm \xi, \bm \psi) = \mathbb{E}_{p(\bm{\theta},\bm{y}|\bm \xi)}[\mathcal{T}_{\bm \psi}(\bm \theta, \bm y)] - \log \mathbb{E}_{p(\bm{\theta})p(\bm y|\bm \xi)}[{\rm clip}(e^{\mathcal{T}_{\bm \psi}(\bm \theta, \bm y)}, e^{-\tau}, e^{\tau})] \label{eq:smile}
\end{equation}
where clip function $\rm clip (u,v,w) = \max(\min(u,w), v)$. The choice of $\tau$ affects the bias-variance trade-off: when $\tau \rightarrow \infty$, $I_{\rm SMILE}$ converges to $I_{\rm MINE}$;  with a smaller $\tau$, the variance is reduced at the cost of increasing bias \cite{song2019understanding}. The improved MI estimation via variance reduction techniques is benefit to the optimization process in Eq.~\eqref{eq:smile} and thus leads to a robust final optimal design $\bm{\xi}^*$.

{\bf Evolutionary strategies} \quad When the gradient of the MI lower bound is inaccessible, a popular zero-order approach for estimating the gradient is Gaussian Smoothing (GS) (or called Evolution Strategies \cite{salimans2017evolution}). The smoothed loss is defined by 
\begin{equation}
      f_{\sigma}(\bm \xi) = \mathbb{E}_{\bm \epsilon \sim \mathcal{N}(0, \mathbf{I}_d)} \left[f(\bm \xi + \sigma \bm u) \right] = (2\pi)^{-n/2} \int_{\mathbb{R}^n}f(\bm \xi + \sigma \bm \epsilon) e ^{-\left\| \bm\epsilon \right\|^2/2 }d\bm \epsilon  \label{eq:es_fun}  
\end{equation}
where $\mathcal{N}(0, \mathbf{I}_d)$ is the $d$-dimensional standard Gaussian distribution, and $\sigma > 0$ is the smoothing radius. The standard GS represents the $\nabla f_\sigma(\bm \xi)$ as an $d$-dimensional integral and estimate it by drawing $M$ random samples $\{\bm \epsilon_i\}_{i=1}^M$
from $\mathcal{N}(0,\mathbf{I}_d)$, i.e., 
\begin{equation} \label{eq:es_grad} 
    \nabla f_{\sigma}(\bm \xi)  = 
    \frac{1}{\sigma}\mathbb{E}_{\bm \epsilon \sim \mathcal{N}(0, \mathbf{I}_d)} \left[f(\bm \xi + \sigma \bm \epsilon)\, \bm \epsilon \right]  \approx \frac{1}{M\sigma}\sum_{i=1}^M f(\bm \xi + \sigma \bm \epsilon_i)\bm \epsilon_i.
\end{equation}
The MC estimator in Eq.~\eqref{eq:es_grad} is usually used as an unbiased estimator of the local gradient $\nabla F(\bm x)$ by exploiting the fact that $\lim_{\sigma \rightarrow 0} \nabla F_{\sigma}(\bm x) = \nabla F(\bm x)$, where the smoothing radius $\sigma$ is often set to a small value. However, the traditional GS tends to a high variance for high dimensional space. 

Several advances in evolution strategies (ES) \cite{choromanski2019complexity,choromanski2018structured,zhang2020scalable,liuself,zhang2021directional} addressed these issues using variance reduction and dimension reduction strategies. More recently, a Guided ES method \cite{maheswaranathan2018guided} is proposed by optimally using surrogate gradient directional along with a random search. Specifically, Guided ES generates a subspace by keeping track of the previous $k$ surrogate gradients during optimization, and leverages this prior information by changing the distribution of $\bm \epsilon_i$ in Eq.~\eqref{eq:es_grad} to $\mathcal{N}(0, \bm \Sigma)$ with $\bm \Sigma = (\alpha/n) \cdot \mathbf{I}_n + (1-\alpha)/k \cdot UU^T$ where $k$ and $n$ are the subspace and parameter dimensions respectively, U denotes an $n \times k$ orthonormal basis for the subspace, and $\alpha$ is a hyperparameter that trades off variance between the subspace and full parameter space. The improved search distribution allows a low-variance estimate of the descent direction $\nabla f_{\sigma}^G(\bm \xi)$.

{\bf Stochastic approximate gradient ascent (SAGA) method} \ We propose a hybrid gradient method by maximizing $I_{\rm SMILE}(\bm \xi, \bm \psi)$ by gradient ascent on NN parameters $\bm \psi$ combined with an approximate gradient $\nabla f_{\sigma}^G(\bm \xi)$ using Guided ES on design parameters $\bm \xi$. Starting from an initial design $\bm \xi_0$ and NN parameters $\bm \psi_0$, the approximate gradient of the MI lower bound $\nabla_{\bm \xi} I_{\rm SMILE}(\bm \xi, \bm \psi)$ at  $\bm \xi_0$ can be estimated by the Guided ES method. With the gradient in hand, we can maximize the MI lower bound with respect to both the NN parameters $\bm \psi$ and the design parameters $\bm \xi$ jointly by gradient ascent optimizer, which is given by Algorithm 1. The proposed method is named by \texttt{SAGABED}, which is a critical contribution to scale the Bayesian experimental design to high dimensional setting such that we can overcome the grand challenge in gradient-free methods, e.g., Bayesian optimization.     

In the following, we discuss some important features of the \texttt{SAGABED} method, specifically for high-dimensional design problems: (1) {\em Unified framework vs. two-stage framework}: Without the requirement of pathwise gradients for implicit models, we utilize the stochastic approximate gradients and construct a unified framework that allows the design process to be performed by a simultaneous optimization with respect to both the variational and design parameters.  The existing two-stage framework that builds a pointwise MI estimator before feeding this estimator to an outer-level optimizer is often computationally intensive; (2) {\em Scalability, portability, and parallelization}: we propose a stochastic approximate gradient ascent procedure that naturally avoids the scalability issue in gradient-free methods. The proposed framework can be easily incorporated with other MI estimators and implicit models because we only need the function value to approximate the gradient based on the guided ES algorithm; (3) {\em Robust estimation with a low variance}: The smoothed MI lower bound used here allows us to perform a robust MI estimation and optimization of experimental design. The resulting low variance of the optimal design and posterior samples enable a more accurate estimate of the model parameters. 

\begin{algorithm}[h!]\label{algo:1}
\small
  \caption{\hspace{-0.1cm}: The \texttt{SAGABED} algorithm}
\begin{algorithmic}[1]
\STATE{\bf Require}: neural network architectures, learning rates $\ell_{\psi}$ and $\ell_{\xi}$, $\tau$ in $I_{\rm SMILE}$, total prior samples $n$, total iterations $T$, implicit model $\mathcal{M}$
\STATE{\bf Process}:
\STATE Initialize a design $\bm \xi_0$ by random sampling
\STATE Initialize neural network parameter $\bm \psi_0$
\FOR{$t=0:T-1$}
\STATE Draw $n$ samples from the prior distribution of the model parameters $\bm \theta$: 
$ \bm \theta^{(1)},...,\bm \theta^{(n)} \sim p(\bm \theta) $
\STATE Compute the data samples $\bm y^{(i)}$, $i=1,...,n$ using the current design $\bm \xi_t$ and a implicit model $\mathcal{M}$
\STATE Evaluate the smoothed MI lower bound $I_{\rm SMILE}$ at the current design $\bm \xi_t$ and network parameters $\bm \psi_t$
\STATE Compute the approximate gradient estimator $\nabla_{\bm \xi} I_{\rm SMILE} (\bm \xi_t, \bm \psi_t)$ using the GES algorithm
\STATE Evaluate the gradient of the $I_{\rm SMILE}$ with respect to the network parameters $\nabla_{\bm \psi} I_{\rm SMILE} (\bm \xi, \bm \psi)$ 
\STATE Update design $\bm \xi_t$ via gradient ascent: \\
$\bm \xi_{t+1} = \bm \xi_{t} + \ell_{\xi} \nabla_{\bm \xi} I_{\rm SMILE} (\bm \xi_t, \bm \psi_t) $
\STATE Update neural network parameters $\bm \psi_t$ via gradient ascent: \\
$\bm \psi_{t+1} = \bm \psi_{t} + \ell_{\psi} \nabla_{\bm \psi} I_{\rm SMILE} (\bm \xi_t, \bm \psi_t) $
\ENDFOR
\end{algorithmic}
\end{algorithm}


After determining the optimal design $\bm{\xi}^*$ by maximizing the MI lower bound, we can obtain an estimate of the posterior $p(\bm \theta | \bm y,  \bm{\xi}^*)$ given the learned neural network $\mathcal{T}_{\bm{\psi}^*}(\bm \theta, \bm y)$ and prior distribution
\begin{equation}
    p(\bm \theta | \bm y, \bm \xi) = {\rm clip}(e^{\mathcal{T}_{\bm \psi}(\bm \theta, \bm y)-1}, e^{-\tau}, e^{\tau}) p(\bm \theta) \label{eq:posterior}
\end{equation}
The relationship in Eq.~\eqref{eq:posterior} allows to easily generate posterior samples $\bm \theta_i \sim p(\bm \theta | \bm y, \bm \xi^*)$ using MCMC algorithm since the posterior density can be quickly evaluated via Eq.~\eqref{eq:posterior}. 

\section{Experiments}
{\bf Noise linear regression} \quad We first demonstrate our proposed method using a classical noisy linear model \cite{kleinegesse2020bayesian}, which is given by $\bm y= \theta_1 \bm 1 + \theta_2 \bm d + \bm \epsilon + \bm \nu$, where $\bm y$ is a response variable, $\bm \theta=[\theta_1, \theta_2]^T$ are model parameters, $\bm \epsilon \sim \mathcal{N}(0,1)$ and $\bm \nu \sim \Gamma(2,2)$ are noise terms. Our target is to make $D$ measurements to estimate $\bm \theta$ by constructing a design vector $\bm d = [d_1,...,d_D]^T$ which consists of individual experimental designs. Four cases $D$=1, 10, 50 and 100 are investigated in this example. We randomly initialize a design $d \in [-10,10]$ and sample 10,000 parameters from a prior distribution $p(\bm \theta) = \mathcal{N}(0, 3^2)$ and we use one layer of 100 and 150 hidden neurons for $D=1$ and $D=10$ respectively. For high dimensional cases $D=50$ and $D=100$, we use 5-layered network with 50 hidden neurons for each layer. These NN architecture settings follow up the guidance in \cite{kleinegesse2020bayesian}. 
\begin{figure}[h!]
    \centering
    \includegraphics[width=0.24\textwidth]{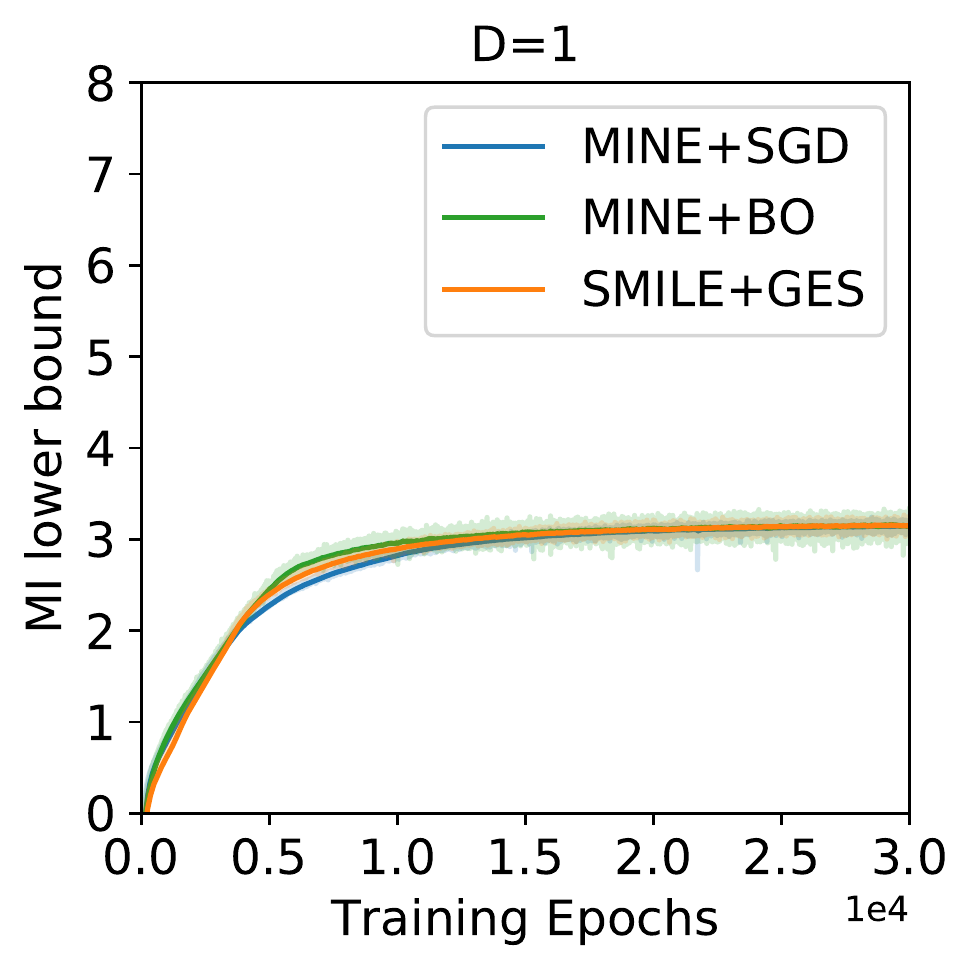}
    \includegraphics[width=0.24\textwidth]{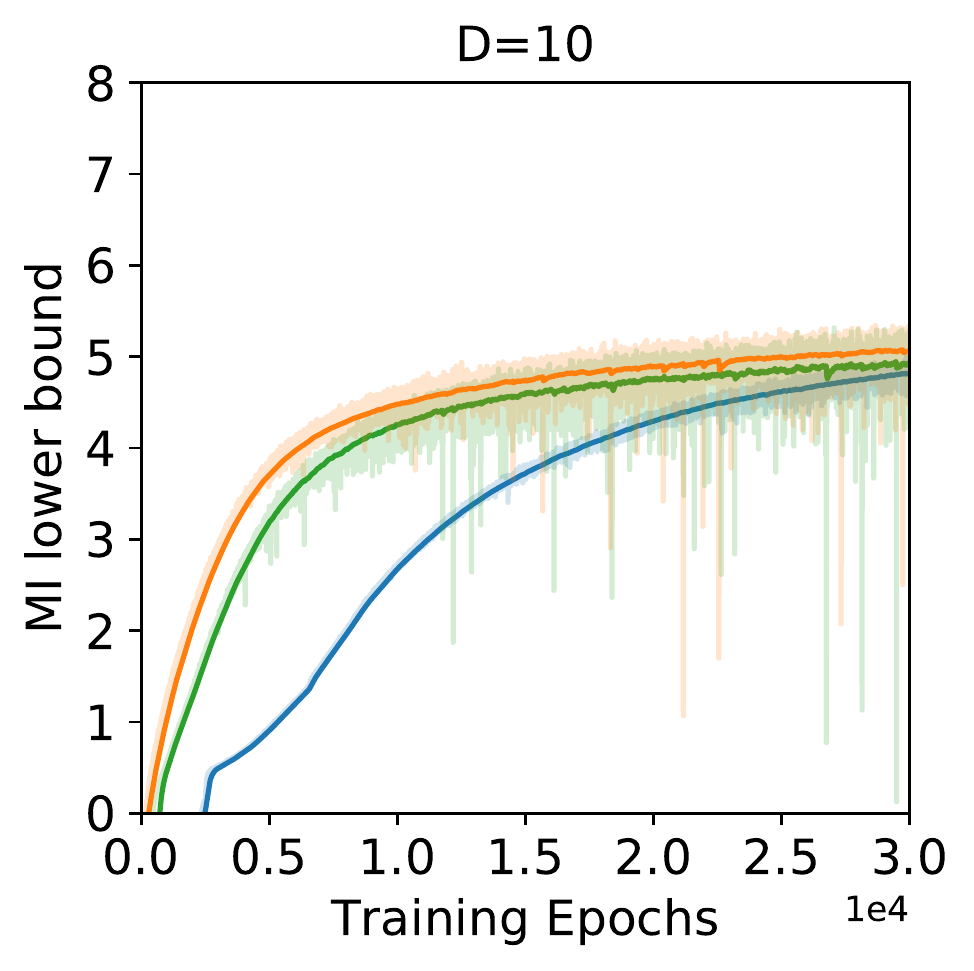}
    \includegraphics[width=0.24\textwidth]{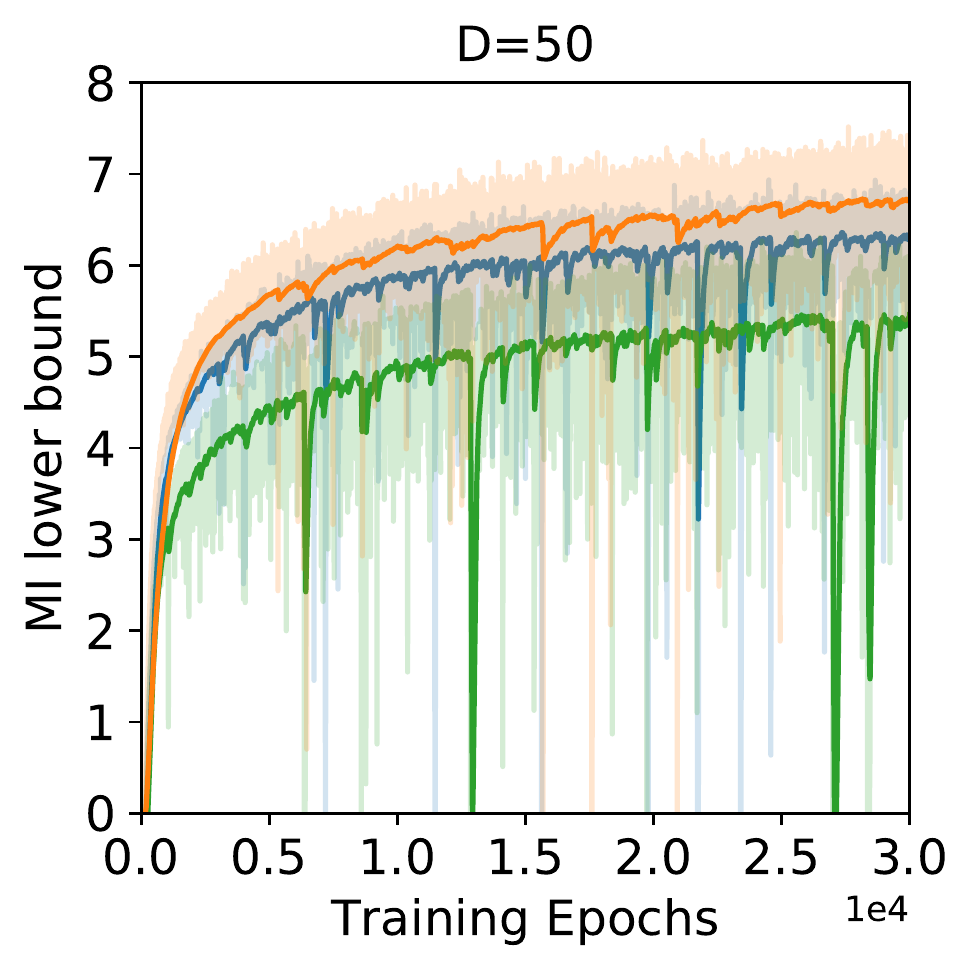}
    \includegraphics[width=0.24\textwidth]{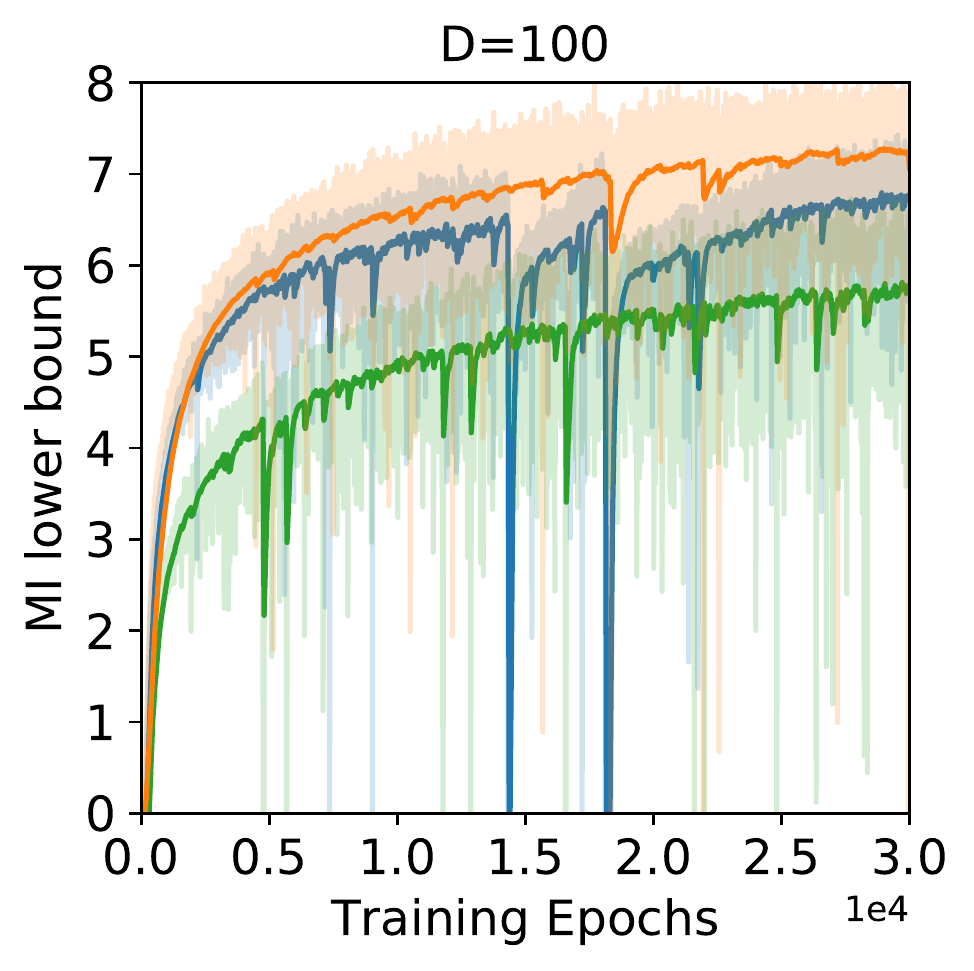}
    \vspace{-0.3cm}
    \caption{MI lower bound as a function of NN training epochs for $D$=1, 10, 50 and 100.}
    \label{fig:noise}
    \vspace{-0.3cm}
\end{figure}

Figure \ref{fig:noise} shows a comparison between the proposed \texttt{SAGABED} method (combined SMILE with Guided ES) and two baselines (MINE with SGA and BO). When $D$=1, three methods perform similarly but our method outperforms the other two baselines on the high dimensional cases. The SGA method is close to our method but shows a unstable training with large variance, and is probably trapped into a local minimal when $D$=10. The performance of the MINE with BO method drops significantly as the dimension increases. The proposed \texttt{SAGABED} method also demonstrates superior performance in estimating the posterior distributions, which is clearly reflected by the smaller variance of the posterior samples in Table \ref{sample-table1}. 

\begin{table}[] 
\footnotesize
\centering
\caption{Estimating mean and standard deviation of the posterior samples of the model parameters $\bm \theta$ using optimal designs $\bm d^*$ and real data observation $\bm y^*$ (use $\bm \theta_{\textup{true}}$ = [1,4] to generate $\bm y^*$)}
\label{sample-table1}
\begin{tabular}{@{}ccccccc@{}}
\toprule
Method      & \multicolumn{2}{c}{D=10}   & \multicolumn{2}{c}{D=50}  & \multicolumn{2}{c}{D=100} \\ \midrule
            & $\hat{\theta_1}$      & $\hat{\theta_2}$      & $\hat{\theta_1}$      & $\hat{\theta_2}$     & $\hat{\theta_1}$      & $\hat{\theta_2}$     \\
MINE + SGA  & 0.51$\pm${\bf0.44} & 2.99$\pm$0.67 & 1.20$\pm$0.18 & 3.79$\pm$0.23  & 0.97$\pm$0.05   & 4.04$\pm$0.04  \\
MINE + BO   & 1.22$\pm$0.58 & 4.93$\pm$0.91 & 0.71$\pm$0.25   & 3.66$\pm$0.40 & 1.35$\pm$0.11   & 4.79$\pm$0.26  \\
SMILE + GES & 0.83$\pm$0.56   & 4.69$\pm${\bf0.58} & 1.11$\pm${\bf0.13}   & 4.25$\pm${\bf0.19}  & 1.02$\pm${\bf0.04}   & 3.98$\pm${\bf0.03}  \\ \bottomrule
\end{tabular}
\vspace{-0.5cm}
\end{table}

\begin{figure}[h!]
    \centering
    \vspace{-0.2cm}
    \includegraphics[width=1.0\textwidth]{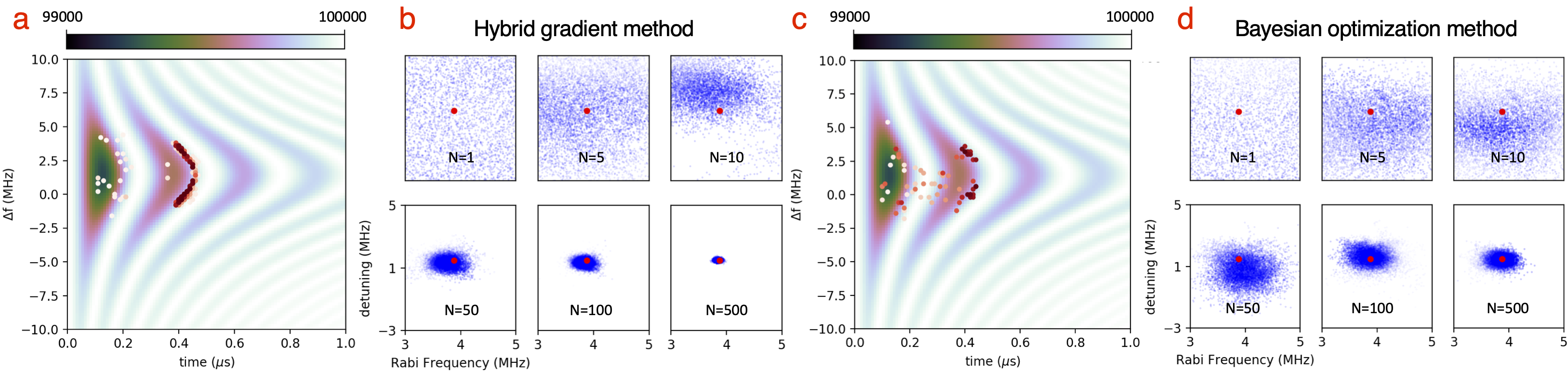}
    \caption{Performance comparison between \texttt{SAGABED} (a-b) and BO (c-d) method on BED for tuning quantum pulse example. The contour images ((a) and (c)) show the model photon counts for optically detected spin manipulation for pulse durations (x-axis) and amounts of detuning from the spin's natural resonance frequency. (b) and (d) display the evolution of the posterior distributions with the number of designed measurements. The red points in (b) and (d) are the true mean photon count. 
    }
    \label{fig:quantum}
    \vspace{-0.2cm}
\end{figure}

{\bf Tuning for quantum control} \quad Typically, the reliable capability to manipulate qbit states is critical to quantum technologies. For instance, radio-frequency pulses can be used to change of state of spin-up and spin-down, and patients benefit from MRI scans that use these approaches. In this example, we use the BED to simulate a tuning process such that we can control the desired duration and frequency of pulses to flip electron spins in a reliable scheme. For the implicit model without access to gradients, we thus compare our proposed \texttt{SAGABED} method with the Bayesian optimization method. As shown in Figure \ref{fig:quantum}, our method outperforms the Bayesian optimization in terms of better measurement designs in Figure \ref{fig:quantum}(a) and a much smaller variance in the posterior distributions of the high-dimensional cases such as $N=$ 50, 100 and 500 in Figure \ref{fig:quantum} (b). 


{\bf Related works}. Foster et al. \cite{foster2019variational} recently proposed to use the MI lower bound for Bayesian optimal experimental design. This study relies on variational approximations to the likelihood and posterior but it is a two-stage approach where the optimal designs were determined by a separate BO. As a result, this approach has a limitation in scaling to high-dimensional design problems. A follow-up study developed by \cite{foster2020unified} aims to address the scalability issue by introducing a unified stochastic gradient-based approach. However, they assumed the models with the tractable likelihood or the gradient approximations are available. In the scope of BED for implicit models, Ao and Li \cite{ao2020approximate} proposed an approximate KLD based BED method for models with intractable likelihoods; Kleinegesse and Gutmann \cite{kleinegesse2019efficient, kleinegesse2020sequential} have recently considered the use of the MI combined with likelihood-free inference by ratio estimation to approximate posterior distributions but this method is often computationally intensive. The authors rectify this in a follow-up study \cite{kleinegesse2020bayesian} that leverages MINE to jointly determine the optimal design and the posterior. However, these methods relied on a outer-level optimizer such as BO are difficult to address the scalability issues.

\section{Conclusion}
In this paper, we develop a hybrid gradient method that leverages recent developments in variational MI estimator and evolution strategies. The proposed \texttt{SAGABED} method incorporating a smoothed MI estimator with a guided ES achieves a unified scalable procedure to simultaneously determine the optimal design and NN parameters for implicit models without a sampling path gradient. The performance is demonstrated by one classical noisy linear model and one scientific quantum control example. The results show that our proposed method outperforms the other two baselines in terms of the MI lower bound estimate and the variance of the posterior samples given the optimal design. 

\section{Acknowledgments}
This work was supported by the U.S. Department of Energy, Office of Science, Office of Advanced Scientific Computing Research, Applied Mathematics program under contract ERKJ352, ERKJ369; and by the Artificial Intelligence Initiative at the Oak Ridge National Laboratory (ORNL). ORNL is operated by UT-Battelle, LLC., for the U.S. Department of Energy under Contract DEAC05-00OR22725.

\section{Broader impact}
Since this work belongs to development of fundamental machine learning algorithms, it does not present any immediately foreseeable societal consequence. However, further development of the proposed method may lead to some potential positive or negative impacts on the society. Some positive impacts include (1) enable scientists and engineers to use their implicit simulator code to solve high dimensional Bayesian experiment design for scientific discovery; (2) help experimental scientist design more efficient experiments to reduce experiment cost. Furthermore, we should be cautious of the consequence of failure of the method which could cause incorrect scientific predictions that may delay the progress of scientific discovery. 

\bibliographystyle{unsrt} 
\bibliography{reference.bib}

\end{document}